\documentclass[journal]{IEEEtran}
\usepackage{xcolor,soul,framed} 
\colorlet{shadecolor}{yellow}
\usepackage[pdftex]{graphicx}
\graphicspath{{../pdf/}{../jpeg/}}
\DeclareGraphicsExtensions{.pdf,.jpeg,.png}
\usepackage[cmex10]{amsmath}
\usepackage{array}
\usepackage{mdwmath}
\usepackage{mdwtab}
\usepackage{eqparbox}
\usepackage{url}
\usepackage{verbatim}
\usepackage{adjustbox}
\usepackage{amsmath}
\usepackage{amssymb}
\usepackage{algorithm}
\usepackage[noend]{algpseudocode}

\usepackage{amsmath}

\usepackage{subcaption}
\DeclareCaptionSubType*[Alph]{table}
\DeclareCaptionLabelFormat{mystyle}{Table~\bothIfFirst{#1}{ ̃}#2}
\usepackage{rotating}

\usepackage{color}
\usepackage{xcolor}
\usepackage{comment}
\usepackage{graphicx}
\usepackage{booktabs}
\usepackage[margin=2.5cm]{geometry}
\usepackage{caption}

\begin{document}
\bstctlcite{IEEEexample:BSTcontrol}
    \title{Safe Deep Q-Network for Autonomous Vehicles at Unsignalized  Intersection}
  \author{Kasra~Mokhtari
      and~Alan~R.~Wagner

  \thanks{K. Mokhtari is with the Department of Mechanical Engineering, The Pennsylvania State University, State College, PA, 16802 USA (e-mail: kbm5402@psu.edu).}
  \thanks{A. R. Wagner is with the Department of Aerospace Engineering, The Pennsylvania State University, State College, PA, 16802 USA (alan.r.wagner@psu.edu).}}
  

\maketitle
\begin{abstract}

We propose a safe DRL approach for autonomous vehicle (AV) navigation through crowds of pedestrians while making a left turn at an unsignalized intersection. Our method uses two long-short term memory (LSTM) models that are trained to generate the perceived state of the environment and the future trajectories of pedestrians given noisy observations of their movement. A future collision prediction algorithm based on the future trajectories of the ego vehicle and pedestrians is used to mask unsafe actions if the system predicts a collision. The performance of our approach is evaluated in two experiments using the high-fidelity CARLA simulation environment. The first experiment tests the performance of our method at intersections that are similar to the training intersection and the second experiment tests our method at intersections with a different topology. For both experiments, our methods do not result in a collision with a pedestrian while still navigating the intersection at a reasonable speed.

\end{abstract}

\IEEEpeerreviewmaketitle
\section{Introduction}
\label{sec1}

\begin{figure}[t]
    \centering
    \includegraphics[scale = 0.9]{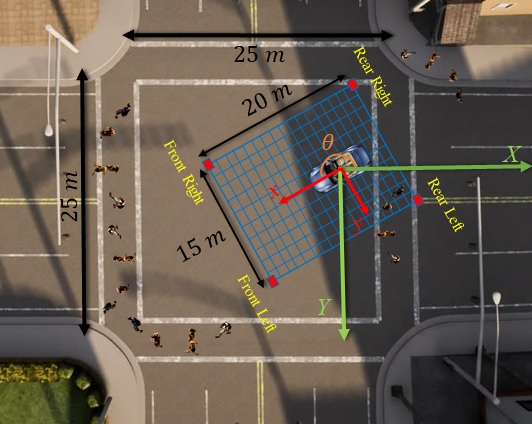}
    \caption{A Bird's eye view of unsignalized intersection is depicted. The global axes are presented with the blue arrows. The region of interest (ROI) is shown by the green grid. For better visualization, the grid is displayed as larger than was used in the simulation (best viewed in color).}
    \label{fig1:intersection}
\end{figure}

\IEEEPARstart{A}{ccording} to the U.S. Federal Highway Administration, approximately 70\% of fatalities due to intersection-related collisions occurred at unsignalized intersections~\cite{coakley2009intersection}. Driving through urban unsignalized intersections without the guidance of centralized traffic lights and traffic signs is a challenging task for an autonomous vehicle (AV). In these scenarios, an AV must autonomously decide when and how to navigate an intersection safely and efficiently, accounting for the intention of pedestrians, cyclists, and other cars. 

Reinforcement learning (RL) combined with deep learning has achieved outstanding success in robotics~\cite{kober2013reinforcement}. Deep reinforcement learning (DRL), in particular, has been used to tackle challenging control problems such as autonomous driving~\cite{tram2018learning, deshpande2019deep}. The application of deep reinforcement learning to autonomous driving mainly focuses on maximizing the long-term reward, which does not prevent the rare occurrence of large negative outcomes (e.g., collisions with pedestrians or other cars). Thus, standard DRL methods that only rely on trial and error searches are not suitable for real-world navigating scenarios where failures can lead to disastrous consequences. Therefore, safety constraints must be enforced when designing AV control mechanisms. 

This paper explores the problem of developing an AV control method that can safely navigate a crowded intersection at a reasonable speed. We are motivated by the observation that special events such as concerts or sports competitions can generate large crowds of pedestrians. These crowds may perambulate through intersections near the event, observing or not observing traffic laws and norms. Moreover, the timing of these crowds can be difficult or impossible to plan around. If we expect ubiquitous autonomous vehicles to navigate our streets, then they will need to be able to manage movement through these crowds without collisions and at a reasonable speed.  

Although, prior research has explored AV navigation in the presence of other vehicles at unsignalized intersections~\cite{isele2018navigating, isele2018safe, liu2020decision}, navigating among crowds of pedestrians, who are the most vulnerable element in the urban environment, is underexplored. Additionally, most of this existing research assumes that the ego vehicle observes the true state of the environment. However, in realistic autonomous driving scenarios, the ego vehicle receives noisy, imperfect observations of the environment. This paper presents a novel collision-free DRL approach to controlling an autonomous vehicle in dynamic, crowded, unsignalized intersections. Moreover, we test our approach in a high-fidelity simulation environment (i.e. CARLA) under the presence of noisy observations. An example of test environment is depicted in Figure~\ref{fig1:intersection}.


The remainder of this paper is organized as follows: Section~\ref{sec2} presents related work, Section~\ref{sec3} describes the Markov decision process (MDP) framework and deep reinforcement learning algorithm (DRL). The proposed safe DRL algorithm implementation is discussed in Section~\ref{sec4}. Section~\ref{sec5} introduces the simulation environment and experiments. Section~\ref{sec6} presents the results from our experiments, and finally Section~\ref{sec7} offers conclusions and directions for future work.

\section{Related Work}
\label{sec2}


\subsection{Autonomous Navigation Around Pedestrians}
\label{sec3.A}
Prior research has studied how to create small ground vehicles capable of navigating through crowds~\cite{kruse2013human}. Discrete sequential game theory has also been used to negotiate collision avoidance between the AV and a pedestrian at an unsignalized intersection~\cite{fox2018should}. Game-theory models are, however, too simplistic to capture complex interactions between AVs and pedestrians. Barbier et al. used a partially observable Markov decision process (POMDP) to model an AV's interactions with pedestrians at intersections~\cite{barbier2018probabilistic}. POMDPs are capable of representing several forms of uncertainty. Although exact solutions to POMDPs are intractable, approximations are possible. Yet, the safety of these approximations when applied to autonomous driving is still a topic of research~\cite{rodriguez1999reinforcement}.

Reinforcement learning (RL) has been been a promising approach for autonomous vehicle navigation around pedestrians. Deep reinforcement learning has also been used to control an autonomous vehicle in the presence of pedestrians. Deshpande et al. demonstrate a system that relies on a Deep Q Network (DQN) allowing the vehicle to cross an intersection in the presence of pedestrians~\cite{deshpande2019deep}. They assume a grid-based state-space representation of the environment and an overly simplified pedestrian model. Unfortunately, the learned policy does not guarantee pedestrian safety.

\subsection{Safe Reinforcement Approaches for AV Navigation}
\label{sec3.A}

Safe reinforcement learning (SRL) attempts to learn policies that maximize long-term reward while ensuring safety constraints during the learning and/or deployment. To ensure safety, two approaches are used: 1) modification of the optimality criterion by discounting the finite/infinite horizon with a safety factor, and 2) modification of the exploration process through the incorporation of external knowledge (e.g., the guidance of a risk metric)~\cite{garcia2015comprehensive}. 

Using traffic rules and short-term predictions, conservative rule-based strategies can constrain the action space of the RL agent in lane changing scenarios~\cite{mukadam2017tactical, mirchevska2018high}. Although rule-based methods are relatively reliable and easily interpreted, they require full knowledge of the environment. POMDPs and SRL are integrated in~\cite{bouton2019safe} to develop efficient policies with probabilistic safety assurances while also including the AV's interactions with pedestrians and other cars. Uncertainty about the other road users' course of action is captured by the transition model and state uncertainty. Although this RL agent guarantees safe navigation of the AV through pedestrians, the computation time grows dramatically with the number of road users in the environment. Therefore, this model is not suitable for real-time autonomous driving applications. Our safe DRL approach is inspired by~\cite{isele2018safe} in which the authors predict and then mask unsafe actions while the AV makes a turn at an unsignalized intersection in presence of other cars using stochastic games.    

\section{Technical Background}
\label{sec3}
\subsection{Reinforcement Learning}
\label{sec3.A}
In a reinforcement learning framework where an agent interacts with the environment over time, at each timestep $t$ the agent takes an action $a_t$ based on a policy $\pi(a|s)$, it receives a scalar reward $r_t$ and transitions into the next state $s_{t+1}$. This continues until the agent reaches a terminal state, at which point it restarts. 

A reinforcement learning task is typically formulated as a Markov decision process (MDP) defined by the 5-tuple $(S, A, P, R, \gamma)$, where $S$ is the set of states, and $A$ is the set of actions that the agent may execute~\cite{sutton2018reinforcement}. MDPs follow the Markov assumption, which is the probability of transitioning to a new state, given the current state and action, is independent of all previous states and actions. The state transition probability $P: S\times A \times S\rightarrow [0,1]$ represents the system dynamics, the reward function $R: S\times A \times S\rightarrow r$ offers the agent a scalar reward for a given timestep, and a discount factor $\gamma \in (0,1]$ discounts the future reward while preventing instability in the case of infinite time horizons. The goal of the agent is to learn an optimal policy $\pi^*$ to maximize the discounted, accumulated reward defined as:
\begin{equation}
R_t=\sum_{k=0}^\infty \gamma^k r_{t+k}.
\end{equation}
To tackle this optimization problem, Q-learning is used.

\subsection{Q-learning}
\label{sec3.B}

Q-learning is one of the most popular reinforcement learning approaches where the agent attempts to learn an optimal state-action value function without prior knowledge of the environment. For a state-action pair $(s,a)$ following a policy $\pi$, the expected reward is described as:
\begin{equation}
Q^{\pi}(s,a) = \mathbf{E}[R_t|s_t=s, a_t=a].
\end{equation}
According to the Bellman equation, the optimal state-value function $Q^{*}(s,a)$ is:  
\begin{equation}
Q^{*}(s,a) = \mathbf{E}[r+\gamma \max_{a_{t+1}}Q^{*}(s_{t+1},a_{t+1})|(s,a)].
\end{equation}
Deep Q Network (DQN) and Double Deep Q Network (DDQN) are recent approaches to Q-learning that use neural networks to approximate $Q^{*}(s,a)$. These methods are described below:
\subsubsection{Deep Q Network (DQN)}
\label{sec3.B.1}
For a DQN, the optimal state-action value function $Q^{*}(s,a)$ is estimated using a neural network to approximate the value $Q^{*}(s,a)\approx Q(s,a,\theta)$, where $\theta$ are the weights of the network~\cite{mnih2015human}. The weights are learned by iteratively minimizing the error between the expected reward and the state-action value predicted by the network. To train the DQN more effectively, replay memory is employed in which the transition sequences are stored in a reply buffer as the 4-tuple of $(s_t, a_t, r_t, s_{t+1})$ at every timestep. The transition includes the current state, the selected action, the corresponding received reward, and the subsequent state. To prevent correlation in the data, a mini-batch of size $N$ is randomly sampled from the reply memory buffer and feed to the network to train the DQN using the loss function: 
\begin{equation}
\label{eq.y_DQN}
y^{DQN}_{t}=r+\gamma \max_{a_{t+1}}Q(s_{t+1},a_{t+1};\theta)
\end{equation}

\begin{equation}
\mathbf{L(\theta)} = \mathbf{E_N}[(y^{DQN}_{t}-Q(s_t,a_t,\theta))^2].
\end{equation}
Although the DQN algorithm has been extensively used in many areas and can usually achieve excellent performance, it suffers from substantial overestimation resulting from a positive bias that is introduced because Q-learning uses the maximum action value as an approximation for the maximum expected action value. To address this issue, the DDQN was developed~\cite{van2015deep}. 

\subsubsection{Double Deep Q Network (DDQN)}
\label{sec3.B.2}

In a DQN, the state-action value predicted by the network is calculated by maximizing over estimated action values, which tends to select overestimated to underestimated values. If the overestimations are not distributed uniformly and not centered at the current state, the estimate may drift from the true value, negatively affecting the performance of DQN~\cite{van2015deep}. Given the state-action target value using equation (\ref{eq.y_DQN}), the max operator uses the same values both to select and to evaluate actions leading to over-optimistic value estimations. To overcome this issue Van Hasselt et al.~\cite{van2015deep} proposed to decouple the action selection and action evaluation tasks using two separate networks with the same architecture. Thus, a DDQN consists of two networks one called the online network which selects the action (greedy policy) and the second one called target network which is used to estimate the Q value. As a result, two state-action value functions with corresponding parameters $\theta$ and $\theta^{'}$ are learned using a mini-batch of size $N$, and the state-action target value for a DDQN is derived as:
\begin{equation}
y^{DDQN}_{t} = r+\gamma Q(s_{t+1},\max_{a}Q(s_{t+1}, a, \theta_t);\theta_{t^{'}}).
\end{equation}
\begin{equation}
\mathbf{L(\theta)} = \mathbf{E_N}[(y^{DDQN}_{t}-Q(s_t,a_t,\theta))^2].
\end{equation}
wherein $\theta^{-}$ denotes the parameter for a target network that shares the exact architecture as the online network whose parameters are referenced by $\theta$. To increase the stability of the learning process the target network updates its parameter periodically using the latest parameters $\theta$. 
\subsection{Prioritized Experience Reply (PER)}
\label{sec3.C}
To train a DQN and DDQN, transitions are uniformly sampled from replay memory, regardless of their significance. However, transitions might vary in their task relevance, and therefore, are given different priorities when training the agent. Prioritized experience replay (PER) assigns different sampling weights to each transition based on the calculated Temporal Difference (TD) error~\cite{schaul2015prioritized}. For a DDQN, the TD error for the transition $i$ is derived as:
\begin{equation}
\delta_{i}= r+\gamma Q(s_{t+1},\max_{a}Q(s_{t+1}, a, \theta_t);\theta_{t^{'}})-Q(s_t,a_t,\theta)
\end{equation}
The probability of sampling transition $i$ is then defined as:
\begin{equation}
P(i) = \frac{p_{i}^{\alpha}}{\sum_{k}{p_{k}^{\alpha}}}
\end{equation}
where $p(i) > 0 $ is the priority of transition $i$, k is the total number of transitions in the replay memory and $\alpha$ determines how much priority is desired (with $\alpha = 0$ corresponding to uniform random sampling). In proportional prioritization, $p(i)=\delta_{i}+\epsilon$, where $\epsilon$ is a small positive constant. Importance sampling is applied to avoid bias in the updated distribution and the importance-sampling (IS) weights are computed as:
\begin{equation}
w(i) = (\frac{1}{N} \times \frac{1}{P(i)})^\beta
\end{equation}
where $N$ is the replay memory size, and $\beta$ determines the compensation degree that fully compensates for the non-uniform probabilities $P(i)$ if $\beta = 1$. In practice, $\beta$ linearly anneals from its initial value $\beta_{0}$ to $1$. Eventually, to guarantee the stability of the network, the weights are normalized by $1/\max_{i}w_i$.  

\subsection{Long Short-Term Memory (LSTM)}
\label{sec3.D}
Our approach uses LSTMs to model observation uncertainty. An LSTM is a type of Recurrent Neural Network (RNN). Recurrent neural networks are a type of artificial neural network particularly suited to temporal or sequential data as their internal state allows information to persist in contrast to a traditional feedforward network. Using input, output, and forget gates, a Long Short Term Memory (LSTM) RNN overcomes limitations such as the vanishing and exploding gradient problems that plague traditional RNNs~\cite{lstm97}.

\subsection{State Uncertainty}
\label{sec3.D}
In the real world, the agent might not receive accurate information about its observations of pedestrians and other items in the world. A partially observable Markov decision process (POMDP) has been used in some research approaches to autonomous driving where the agent receives imperfect observations about the environment due to sensor error~\cite{barbier2018probabilistic}. In a POMDP, the agent represents its knowledge of the environment with a belief state $b: S \times O \rightarrow  [0,1]$ such that $O$ is an observation space and $b(s)$ is the probability of being at state $s$~\cite{rodriguez1999reinforcement}. The agent receives an observation $o$ and updates its belief at every timestep. Since solving decision-making problems under this formulation is generally interactable, an approximation approach such as QMDP is typically used~\cite{karkus2017qmdp}. However, this approach is still intractable for our crowded intersection problem because our problem involves an infinite number of states.

In this paper, to overcome this issue, the belief update model is trained to handle up to a certain level of perceptual noise and the output of this model is a perceived state $s^p$, given the imperfect observation $o$ at each timestep. We therefore substitute $s^p$ for $s$ into the equations from Section \ref{sec3.B}. The decision-making problem can then be modeled using an MDP. The experiments presented below demonstrate that this approach suffices to make our algorithm robust to state uncertainty. 


\section{Methodology}
\label{sec4}

\begin{figure*}[t]
    \centering
    \includegraphics[scale = 0.43]{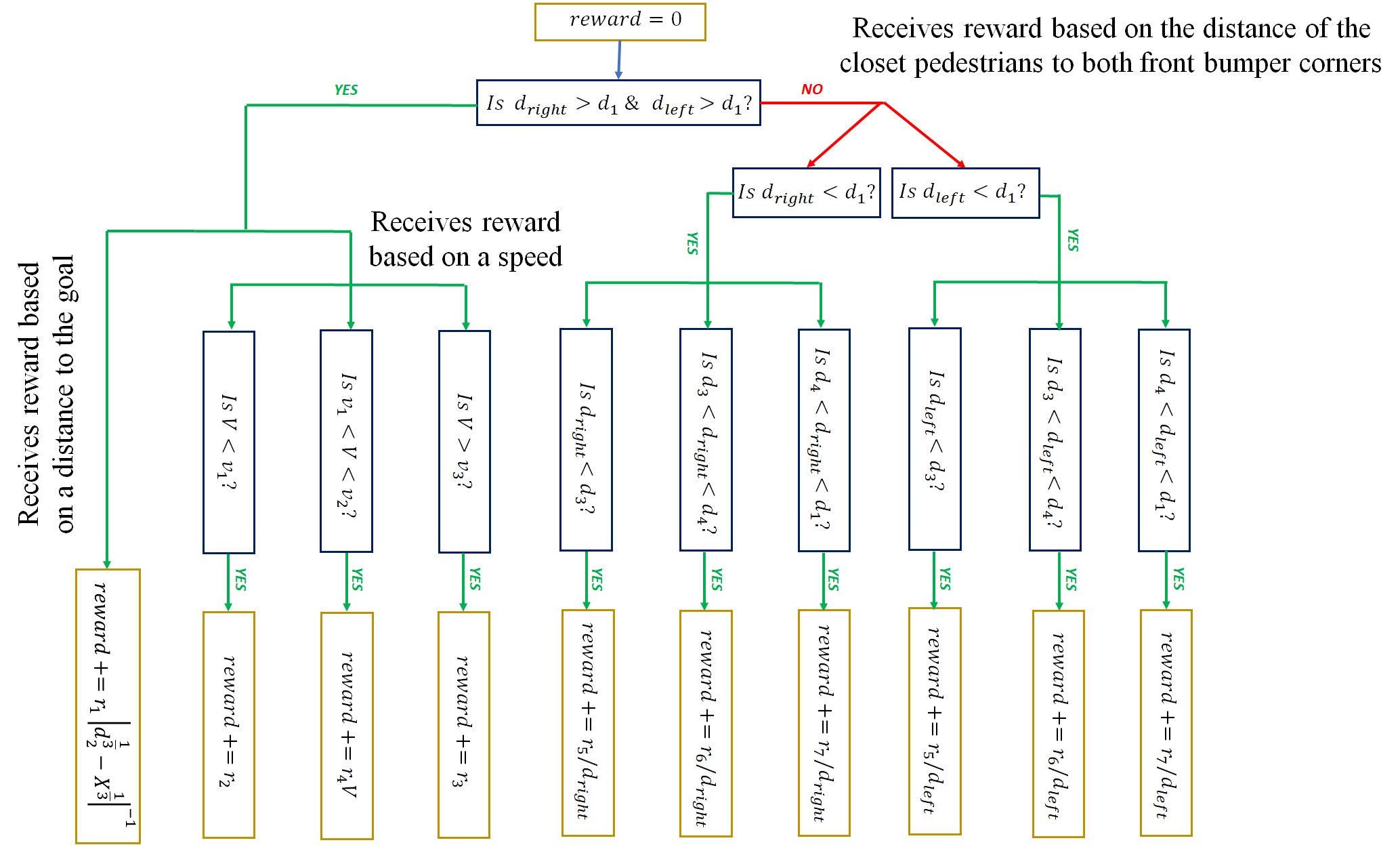}
    \caption{Reward function flow chart.}
    \label{fig2:rewardfunction}
\end{figure*}

We investigate the problem of autonomous vehicle navigation at an unsignalized four-way intersection densely populated with pedestrians randomly moving along crosswalks. The ego vehicle's goal is to make a left-turn in a reasonable amount of time without colliding with pedestrians or violating the speed limit. 

This problem is modeled as an MDP: 
\subsubsection{The State Space}
\label{sec4.B.1}
A multi-layered (3-D) grid is used to represent the environment's state space~\cite{deshpande2019deep}. The resulting 3-D tensor has three dimensions with each corresponding to a particular layer. The region of interest (ROI) of the environment around the ego vehicle is a rectangle with the length $L$ and the width $W$ and is 
discretized into multiple grids each with the grid discretization $l \times w$ as shown in Figure~\ref{fig1:intersection}. The vertices of the ROI are selected with respect to the ego vehicle's center of gravity which always lies in the cell with an index ($\frac{4L}{5l}$,$\frac{W}{2w}$). We assume that the length and the width of the ego vehicle are $4.5m$ and $2.0m$, respectively. The ROI's parameters were $L=20m$, $W=15m$, $l=0.25m$ and $w=0.25m$. As a result, each 2-D grid is an $80 \times 60$ matrix. The ego vehicle's center of gravity occupies cell $(64,15)$ of the first 2-D grid. The ROI's orientation remains the same with respect to ego vehicle direction. Therefore, given the ego vehicle location $(EV_x, EV_y)$ and ego vehicle direction $EV_\theta$, the Cartesian coordinates of the ROI vertices are shown in Table~\ref{tab1:coordinates}. 

\begin{table}[ht]
\centering
\caption{ROI's Cartesian coordinates}
\label{tab1:coordinates}
\begin{tabular}[t]{|c|c|}
\hline
Vertices & Cartesian Coordinates \\
\hline
Rear Right x  & $EV_x+ \frac{L\times cos(\theta)}{5}+\frac{W \times sin(\theta)}{2}$\\ 
Rear Right y  & $EV_y+ \frac{L\times sin(\theta)}{5}-\frac{W \times cos(\theta)}{2}$\\ 
Rear Left x   & $EV_x+ \frac{L\times cos(\theta)}{5}-\frac{W \times sin(\theta)}{2}$\\ 
Rear Left y   & $EV_y+ \frac{L\times sin(\theta)}{5}+\frac{W \times cos(\theta)}{2}$\\ 
Front Right x & $EV_x- \frac{4L\times cos(\theta)}{5}+\frac{W \times sin(\theta)}{2}$\\ 
Front Right y & $EV_y- \frac{4L\times sin(\theta)}{5}-\frac{W \times cos(\theta)}{2}$\\ 
Front Left x  & $EV_x - \frac{L\times cos(\theta)}{5}-\frac{W \times sin(\theta)}{2}$\\ 
Front Left y  & $EV_y - \frac{L\times sin(\theta)}{5}+\frac{W \times cos(\theta)}{2}$\\ 
\hline
\end{tabular}
\end{table}

At each timestep, the 3-D tensor state representation $s^p$ contains three 2-D layers where the first 2-D layer represents the cells that are occupied by the ego vehicle and surrounding pedestrians, relative speed ($m/s$) and relative heading direction ($degrees$) of the corresponding pedestrians with respect to the ego vehicle are stored in the second and the third tensor layers, respectively. 

It is assumed that the pedestrian information collected by the sensors (such as LIDAR) includes some perception error $e$ added to a perfect observation $s$ for each timestep $t$. The belief updater LSTM model is used to calculate the perceived state $s^p$ given the imperfect observation at every timestep $t$ (discussed in Section~\ref{sec5.C}). 

\subsubsection{Action Space}
\label{sec4.B.2}

We control the ego vehicle by adjusting the throttle value $a \in [-1,1]$ and selecting actions from the space of action presented in Table~\ref{tab2:action space}.

\begin{table}[ht]
\centering
\caption{Action Space}
\label{tab2:action space}
\begin{tabular}[t]{|c|c|c|}
\hline
Action & Throttle Value & Description \\
\hline
$a_{0}$  & $-1.0$ & $full_brake$\\
$a_{1}$  & $-0.4$ & $decelerate$\\
$a_{2}$  & $+0.2$ & $accelerate_{1}$\\
$a_{3}$  & $+1.0$ & $accelerate_{2}$\\ 
\hline
\end{tabular}
\end{table}

\begin{table}[ht]
\centering
\caption{Reward function hyperparameters}
\label{tab3:rewardfunction}
\begin{tabular}[t]{|c|c|c|c|}
\hline
Hyperprameter & Value & Hyperprameter & Value \\
\hline
$d_1$ & $7~m$ &$r_2$ & $0.005$\\ 
$d_2$ & $25~m$&$r_3$ & $-0.25$ \\ 
$d_3$ & $1~m$ &$r_4$ & $0.2$\\ 
$d_4$ & $2~m$ &$r_5$ & $-0.5$\\ 
$v_1$ & $1.5~m/s$& $r_6$ & $1.5$\\
$v_2$ & $10~m/s$& $r_7$ & $-0.25$\\
$r_1$ & $0.005$ & &\\
\hline
\end{tabular}
\end{table}

\subsubsection{Reward Function}
\label{sec4.B.3}

The goal of the ego vehicle is to make a left-turn in a reasonable amount of time without colliding with a pedestrian or speeding (max speed of $10~m/s$). We set a time limit for the turn to 45 seconds. To accomplish this, we designed a conditional reward function inspired by~\cite{everett2019collision} where the switching criteria is based on the distance of the ego vehicle's right and left bumpers to the closest pedestrians denoted $d_{right}$ and $d_{left}$, respectively. Therefore, at each timestep, if both distances are greater than a threshold $d_{1}$ from Table ~\ref{tab3:rewardfunction}, then the reward is based on the vehicle's distance to the target position ($X_{t}$) and its velocity ($V$). Otherwise, the reward is based on the distances $d_{right}$ and $d_{left}$ as depicted in Figure~\ref{fig2:rewardfunction}. The hyperparameters of the reward function are shown in Table~\ref{tab3:rewardfunction}. Since the reward function is designed such that the agent receives the highest reward at the end of the navigation task, it is not necessary to assign excessive reward when the agent reaches the goal. The ego vehicle reaches a terminal state when it completes the left-turn, collides with a pedestrian, or it runs out of time, and then the episode restarts. The total amount of reward that the ego vehicle receives for every episode is the summation of all $reward$ over all of the timesteps.
Using deepreinforcement  learning  and  short-term  prediction,  thisarchitecture  is  designed  to  generate  safe  AV  navigationin  crowded  intersections  in  spite  of  noisy  sensor  obser-vations.

\section{System Architecture}
\label{sec5}
\begin{figure*}[t]
    \centering
    \includegraphics[scale = 0.55]{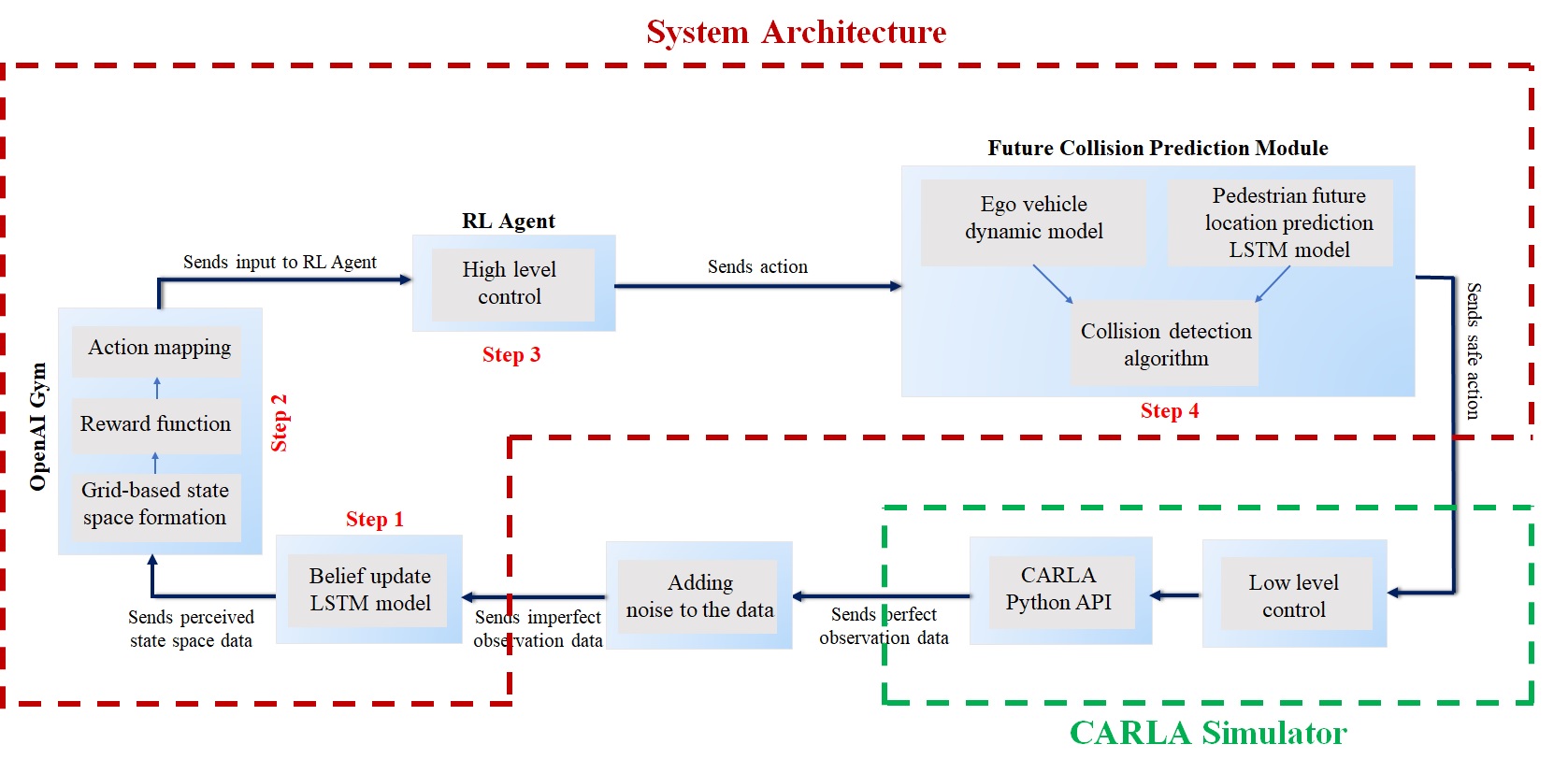}
    \caption{Control architecture for navigating crowded intersections.}
    \label{fig3:Decision-making}
\end{figure*}
In this section, we discuss the implementation details for our approach to navigating crowded intersections in CARLA. Our Control architecture for navigating crowded intersections is shown in Figure~\ref{fig3:Decision-making}. Using deep reinforcement learning and short-term prediction, this architecture is designed to generate safe AV navigation in crowded intersections in spite of noisy sensor observations. 

The noisy observations of the ego vehicle are sent to a belief update LSTM model (Figure~\ref{fig3:Decision-making} step 1) to generate the perceived state, $s^p$. The belief update model is trained to handle up to a certain level of perceptual noise. The perceived state is then fed into the interface component of the OpenAI Gym toolkit~\cite{brockman2016openai} (Figure~\ref{fig3:Decision-making} step 2), which is an open-source library for developing reinforcement learning algorithms. At every timestep, this module calculates the grid-based state-space representation, the assigned rewards to the ego vehicle, and action mapping (as discussed in Section~\ref{sec4}). The high-level control action (a throttle value) for the next timestep is generated by the RL agent and is sent to a future collision prediction module. This module masks the unsafe action if the system predicts a future collision with pedestrians. 

To give the ego vehicle sufficient time to react to a likely collision, the collision prediction is conducted over a fixed simulated time window ($0.5~sec$ from the current moment) with the same timestep taken by the simulator ($1/15~sec)$, called a virtual timestep. During this time window, it is assumed that the action executed by the ego vehicle remains the same for each virtual timestep. Given the ego vehicle dynamic model and the high-level control action, the future trajectory of the ego vehicle is calculated for each virtual timestep. The location of the pedestrians at the end of the simulated time window is computed by a pedestrian future location prediction LSTM model. The trajectory of each pedestrian is derived by calculating a straight line between the start point and the endpoint. Assuming a constant velocity for each pedestrian during this time horizon, the future location of the pedestrian for every virtual timestep is obtained by interpolating along the corresponding trajectory. As a result, at each virtual timestep, the future trajectory of the ego vehicle and the pedestrians are compared, and if the minimum distance of these two trajectories is less than a predefine threshold of $0.5m$, the corresponding high-level action must be masked due to the high probability of a collision. If the high-level actions generated by the RL agent is masked, the high-level control action $a_0=-1$ (as discussed in Section \ref{sec4.B.2}) is sent to CARLA instead. To complete the control loop, the low-level control action is executed by CARLA~\cite{dosovitskiy2017carla}, and the ego vehicle and pedestrian data such as their location, velocity, and heading direction is generated by CARLA API.

\section{Experiments}
 \label{sec8}
 
At the beginning of every episode, the ego vehicle and a random number of pedestrians (between five and thirty) were spawned at the four-way unsignalized intersection in CARLA as shown in Figure~\ref{fig1:intersection}. To ensure that the environment around the ego vehicle always remains crowded, five more pedestrians were spawned every ten seconds after the start of the episode. The pedestrians moved to random destination points with a random velocity between $0.2~m/s$ and $1.8~m/s$. A trajectory was defined for the ego vehicle by automatically placing a hundred waypoints between the start point to the endpoint. 

The ego vehicle and pedestrian data generated by the CARLA API are fed into a noise generator module to create more imperfect observations by adding three types of Gaussian distribution noise $\epsilon=\mathcal{N}(0, \sigma^{2})$ to the actual pedestrian data. The standard deviations of the added noises are $1~m$, $0.2~m/s$, and $10~degrees$ for the pedestrian's location, velocity, and heading direction, respectively. 

The imperfect observations are then sent to a belief update LSTM model to obtain the perceived state. The goal of the ego vehicle was to make a left turn in a reasonable amount of time while avoiding collision with any pedestrian and not violating the speed limit. Forty-five seconds were allotted for each episode. The speed limit for the ego vehicle was $10~m/s$. At each timestep, the action generated by the network was executed in the simulator and the network was trained. The ego vehicle reached the terminal state when it either completes the left-turn, collides with a pedestrian, or runs out of time, at which point the episode restarts. In the following sections, first, the training process of the different components of our system architecture are presented. Then, the empirical experiments of our system are described.   

\subsection{Ego Vehicle Dynamic Model}
\label{sec5.B}
Since the ego vehicle's deterministic dynamics model is not provided by CARLA, we learn it using the ego vehicle dynamics data generated in the simulation. To this end, we use a simple fully connected neural network as a function approximator for the ego vehicle dynamics model, which consists of 4 fully connected layers with 32, 32, 16, 3 units in the first, second, third and fourth layers, respectively. These layers use a ReLU activation function. The input to the model is a $\mathbb{R}^{5\times1}$ vector, where the rows are the low-level control action, the ego vehicle's x and y coordinates, the speed, and the steering wheel value of the current timestep. The output is a $ \mathbb{R}^{3\times1}$ vector, where the rows are the predicted ego vehicle x and y coordinates and the predicted speed for the next timestep. 

We generate the ego vehicle data for 2000 episodes of 100 timesteps of $1/15~sec$ in CARLA. For each episode, the ego vehicle is spawned at a random intersection, and a random low-level action is applied to the ego vehicle at each timestep. The executed low-level action, the ego vehicle x and y coordinates, the speed, and the steering wheel value are reported for every timestep. In order to facilitate training, we preprocess the training data by normalizing and then scaling the data between the range of 0 and 1. Random sub-sequences are then sampled from the data to form the training set. To train the network, we use the Adam optimizer with a learning rate of 0.0001 and the loss function called MSE. During the test phase, we use the mean and standard deviation of the training set to normalize the test data to maintain consistency with the input used to train the network, we then re-scale it to fit in the range of 0 to 1.  

\subsection{Belief Update LSTM Model}
\label{sec5.C}
To capture the state uncertainty related to the pedestrian's data, we propose a belief update LSTM model with the purpose of managing more realistic perception error in the planning algorithm. The belief update is an LSTM model, where the hidden states are responsible for keeping track of the history of the imperfect observations. Using the LSTM model, the perceived true states for each individual pedestrian is calculated separately. Therefore, for each existing pedestrian in the environment, the input to the LSTM model is a $ \mathbb{R}^{3\times4}$ vector, in which the columns are the pedestrian x and y coordinates, velocity, and heading direction, and the rows are the corresponding observations for the timesteps $t-2\Delta t$, $t-\Delta t$, and $t$. The output is the perceived state $\mathbb{R}^{4\times1}$ vector, where the rows are the pedestrian x and y coordinates, velocity, and heading direction for the current timestep $t$ in CARLA. The network architecture includes an LSTM layer consisting of 32 hidden units which was empirically determined to be sufficient. A fully connected layer is then used to obtain an output vector of size 4. The LSTM uses the hyperbolic tangent activation function whereas the dense layer uses a ReLU.

We used CARLA to collect pedestrian data. We generate pedestrian data for 2000 episodes of 400 timesteps of $1/15~sec$. For each episode, one pedestrian is randomly spawned in the environment and the pedestrian x and y coordinates, velocity, and heading direction at every timestep are stored. To create dataset three Gaussian distribution noises $\epsilon=\mathcal{N}(0, \sigma^{2})$ are added to the data, which the standard deviations are $1~m$, $0.2~m/s$, and $10~degrees$ for the pedestrians location, velocity, and heading direction, respectively. To train the network, the features are selected from the imperfect observation dataset while the labels are chosen from the true observation dataset. Similarly to the ego vehicle dynamics model, we normalized the data, used the Adam optimizer with a learning rate of 0.0001, and the MSE loss function. The results in Section \ref{sec5} demonstrate that our LSTM model is robust to perception errors.

\subsection{Pedestrian Future Location Prediction LSTM Model}
\label{sec5.D}

The purpose of this LSTM model is to predict the future location of a pedestrian for the end of the simulated time window ($\sim 0.5$ sec from the current moment) given the history of the corresponding imperfect observations. The prediction task is conducted for each individual pedestrian separately. For each existing pedestrian in the environment, the input to this LSTM model is identical to the input of the belief update model as discussed in Section~\ref{sec5.C}. However, the output is a $\mathbb{R}^{2\times1}$ vector, where the rows are the predicted future x and y coordinates for each pedestrian at the end of the virtual time horizon (as discussed in Section~\ref{sec5}). The network architecture includes an LSTM layer consisting of 32 hidden units. A fully connected layer is then used to derive an output vector of size 2. The LSTM uses the hyperbolic tangent activation function while the dense layer uses a ReLU.

We use the pedestrian data generated in Section~\ref{sec5.B} to train the network. Similarly, the features are chosen from the imperfect observation dataset whereas the labels are selected from the true observation dataset. The data are normalized, and we used the Adam optimizer with a learning rate of 0.0001, and the MSE loss function. 

\subsection{Reinforcement Learning Agent Architecture}
\label{sec5.E}

\begin{figure}[t]
    \centering
    \includegraphics[scale = 0.55]{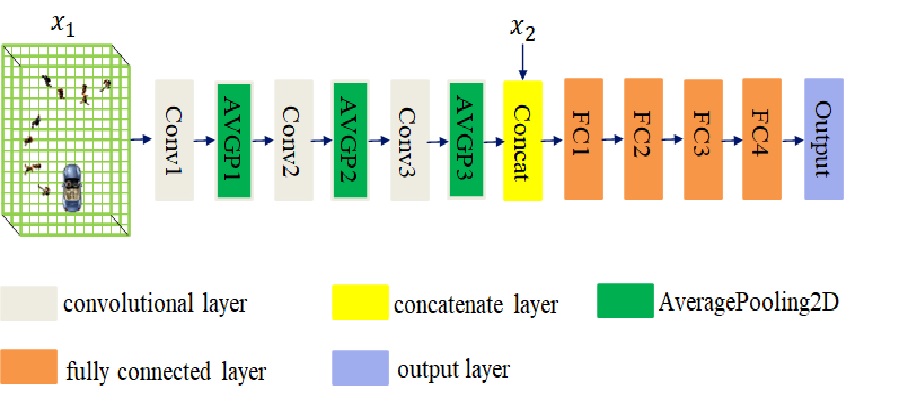}
    \caption{Deep Q-network architecture.}
    \label{fig4:DQN}
\end{figure}

The RL agent combines the DDQN formulation and PER implementation as discussed in Section~\ref{sec3.C}. In the DDQN, the action selection and action evaluation tasks are handled with two separate networks with the same architecture, the online network and the target network. The network architecture is shown in Figure~\ref{fig4:DQN}, and the parameters of the networks are shown in Table~\ref{tab3:DQN}. The input $x_1$ to the agent is the 3-D tenor discussed in Section~\ref{sec4.B.1} and the input $x_2$ is the ego vehicle velocity. Three convolutional layers followed by a ReLU activation function and an Averagepooling2D layer are used to extract the low-level features of the 3D-tensor. The output of the convolutional layers is then fed into a flattened layer and is concatenated with the ego vehicle velocity. The new vector is propagated through four fully connected layers. The target network parameters are updated after 5000 timesteps using the latest parameters of the online network, and the RL agent is then trained as discussed in Section~\ref{sec3.B.2}.

\begin{table}[ht]
\centering
\caption{Parameters of the RL agent architecture.}
\label{tab3:DQN}
\scalebox{0.85}{
\begin{tabular}[t]{|c|c|c|c|c|c|c|c|} 
\cline{1-6}
Layer & Activation Function & Filters & Units & Kernel Size & stride \\
\cline{1-6}
Conv1 & ReLU & $64$ & - & $3\times 3$ & $1 \times 1$  \\
AVGP1 & - & $1$ & - & $5\times 5$ & $3 \times 3$  \\
Conv2 & ReLU & $64$ & - & $3\times 3$ & $1 \times 1$  \\
AVGP2 & - & $1$ & - & $5\times 5$ & $3 \times 3$  \\
Conv3 & ReLU & $64$ & - & $3\times 3$ & $1 \times 1$  \\
AVGP3 & - & $1$ & - & $5\times 5$ & $3 \times 3$  \\
FC1 & ReLU & - & 512 & - & -  \\
FC2 & ReLU & - & 256 & - & -  \\
FC3 & ReLU & - & 64 & - & -  \\
FC4 & ReLU & - & 4 & - & -  \\
\cline{1-6}
\end{tabular}}
\end{table}

To generate training data, the four-way unsignalized intersection left turn scenario was constructed and simulated in CARLA. The simulator ran on synchronous mode (15 FPS) indicating a constant time interval between each timestep taken by the agent in the simulation. The RL agent was trained for 500 episodes. Each timestep generated a transition and the experience replay memory buffer stored up to 10,000 transitions. The learning process began when the experience replay memory reached a threshold of 750 transitions. The agent used a stochastic sampling method (Eq. 9) to choose a mini-batch of size 32 transitions from the replay memory to update the parameters of the local network using the RMSprop optimization method with a learning rate of 0.00025. 

During the training process, a discount factor of 0.95 was applied to discount future rewards. The $\epsilon$-greedy approach was followed in order to allow the networks to explore, a randomly chosen action at each timestep with a probability of $\epsilon$. Otherwise the action selected by the network was used. The value $\epsilon$ was initialized to 1. It annealed to a minimum value of $0.05$ using the decay value of $0.99$ in order to limit the exploratory behavior towards the end of the training process. For the DDQN/PER method, the prioritization rate $\alpha$ was set to $0.6$ and the compensation rate $\beta$ was initialized to $0.4$ linearly annealing to 1.     

\subsection{Scenario}
\label{sec5.F}

To evaluate the effectiveness of belief update LSTM model and collision prediction module to safely navigate among pedestrians, five different agents are defined:
\begin{itemize}
  \item \textbf{Rule-based agent:} This agent is provided by CARLA which executes a policy based on a hand-engineered strategy for navigating around pedestrians. The policy mainly relies on a time to collision approach to decide when to cross the intersections. 
  \item \textbf{RL agent:} The RL agent uses neither the belief update LSTM model nor the future collision prediction module in its decision-making process. Therefore, this agent relies only on the RL module to execute the action as shown in Figure~\ref{fig3:Decision-making}.
  \item \textbf{Belief update agent:} This agent uses only the belief update LSTM model to obtain the perceived state of the environment given the imperfect observations. The action selected by the RL module is sent directly to the simulator without being fed into the future collision detection module. 
  \item \textbf{Future Collision detector agent:} This agent uses the future collision detection module to mask the unsafe action nominated by the RL agent to avoid a potential collision with pedestrians. Moreover, the imperfect observations of the environment are also directly sent to OpenAI Gym without using the belief update LSTM model.
  \item \textbf{SRL agent:} This agent combines the belief update LSTM model and the future collision detection module along with the RL module as shown in Figure~\ref{fig3:Decision-making}. If the action given by the RL module is within a set of safe actions, then the action is executed. Otherwise, it executes the safest action $a_0=-1$.
\end{itemize}

Given the corresponding decision-making strategy, all of the agents were trained for 500 episodes as discussed in Section\ref{sec5.E}. The testing performance of the agents is compared in two experiments. Both experiments were conducted for 200 episodes, used the same definition of the terminal state, the same training process and method of spawning pedestrians. In the first experiment, these agents were tested on exactly the same unsignalized intersection on which the networks were trained. For the second experiment, the networks were tested on an unseen unsignalized intersection with different topology. The networks were trained on a $22~m \times 22~m$ four-way unsignalized intersection while the unseen intersection was a $26~m \times 20~m$ three-way unsignalized intersection. For each experiment, the following metrics were used to evaluate each methods' performance: the percentage of successful episodes, collision episodes, percentage of speed limit violations, percentage of vehicle times out, average intersection crossing time (sec), average interaction crossing speed (m/s), and the average distance of ego vehicle from the closest pedestrian (m).

\section{Results and Discussions}
\label{sec6}

\begin{table*}
\caption{Comparison of test performance results}
\label{tab5:testresults}
\bigskip
\scalebox{0.85}{
\begin{tabular}{@{}*{8}{c}@{}}
\multicolumn{8}{@{}l}{\em(a) The intersection that the agents are trained on}\\
\toprule
Agent & Successful & Collisions & Out of Time & Speed violation & Average intersection & Average intersection & Average Distance from\\
& episodes& episodes & episodes & episodes &  crossing time (sec) & crossing speed (m/s) & the closet pedestrian(m)  \\
\midrule
Rule-based & 38\% &62\%& 0\% & 0\% & 17.57 & 4.70 & 4.8\\                                                     
RL & 84\% &6\%& 10\%  & 0\% &  24.55 & 3.44 & 6.84\\                                                               
Belief Update &88\% &4\%& 8\% & 0\% &  25.32 & 3.38 & 6.81\\                                                         
Future Collision Detector & 88\% &0\%& 12\% & 0\% &  30.14 & 3.11 & 7.02\\                          
SRL & \textbf{92}\% & 0\%& 8\% & 0\% &  28.12 & 3.25 & 7.19\\                          

\bottomrule
\end{tabular}}

\bigskip
\scalebox{0.85}{
\begin{tabular}{@{}*{8}{c}@{}}
\multicolumn{8}{@{}l}{\em(b) An unseen intersection with different topology}\\
\toprule
Agent & Successful & Collisions & Out of Time & Speed violation & Average intersection & Average intersection & Average Distance from\\
& episodes& episodes &  episodes & episodes &  crossing time (sec) & crossing speed (m/s) & the closet pedestrian(m)  \\
\midrule
Rule-based & 36\% & 64\%& 0\% & 0\% & 18.38 & 4.76 & 4.62\\                                                
RL & 81\% & 7\%& 12\% & 0\% &  25.18 & 3.48 & 6.88\\                                                               
Belief Update & 86\% &4\%& 10\% & 0\% &  25.74 & 3.41 & 6.85\\                                                         
Future Collision Detector &88\% &0\%& 12\% & 0\% &  30.66 & 3.15 & 7.10\\                          
SRL & \textbf{91}\% & 0\%& 9\%  & 0\% &  28.61 & 3.27 & 7.21\\                          

\bottomrule

\end{tabular}}
\end{table*}

The results for the two experiments are presented in Table~\ref{tab5:testresults}. The first column represents the successful episodes which are the episodes that the agent neither had a collision with pedestrians, nor ran out of time, nor violated the speed limit. Among all the agents, the SRL agent shows the greatest successful episodes with $92\%$, $91\%$ for the first experiment and the second experiment, respectively. 

Although the average speed and the average intersection crossing time for the rule-based agent outperforms our methods, 62\% and 64\% of the episodes result in collisions with pedestrians for experiment one and experiment two, respectively. As a result, the rule-based method is not a viable solution to this problem. For the first experiment, compared to the rule-based approach, our methods clearly show better performance with $6\%$, $4\%$, $0\%$, $0\%$ collision episodes for RL, belief update, future collision detector, and SRL, respectively. Qualitatively, we see that these methods learn to decrease their speed and wait for pedestrians to cross in order to avoid a collision with a pedestrian. These methods are also more successful at completing the navigation task than rule-based methods. We therefore conclude that our methods are safer by a great margin than the standard rule-based method.  

Considering the performance of the RL agent, it receives a noisy observation of the environment which eventually leads to bad decisions ($\approx6-7\%$ of collisions). Additionally, the RL and the belief update agents that do not rely on the future collision detection module have more collisions than the future collision detector and the SRL agents. The future collision detector and SRL agents have 0\% collisions with pedestrians in both scenarios, an important prerequisite for autonomous driving control. These results confirm the effectiveness of the future collision detector module in masking the unsafe actions generated by the RL module. Compared to the future collision detector agent, since the SRL approach also employs the belief update LSTM model in order to obtain the perceived state of the environment in the case of imperfect observations, it has the faster average intersection crossing time ($28.12 s$ for experiment one and $28.61 s$ for experiment two), the greater average speed ($3.25m/s$ for experiment one and $3.27m/s$ for experiment two), and the greater distance from the closest pedestrian ($7.19m$ for experiment one and $7.21m$ for experiment two) as depicted in Table~\ref{tab5:testresults}. The results demonstrate that the belief update module plays an important role in making the algorithm robust to perception noise, and that using a pure reinforcement learning technique (RL agent) does not guarantee safe navigation around pedestrians in the presence of noisy observation. Overall, the SRL and future collision detector methods are capable of safely navigating through crowds and following the traffic rules for different unsignalized intersections after being trained on one topology and then tested on another topology.

Given the update belief LSTM model and future collision detector module, employing the 3D state space representation of the environment and our innovative conditional reward function combined with the DDQN/PER training methods, to the best of our knowledge, we have developed the first safe reinforcement learning-based decision-making process for an AV tested in a high-fidelity simulator that is not only safe ($0\%$ collision-free episodes) but also capable of successfully navigating at unsignalized intersections in a pedestrian-rich urban environment in the presence of noisy observation regardless of the intersection topology. Moreover, due to the small computation time ($\approx0.1~sec$) required by the networks, we believe that these methods can be used in real-time.

\section{Conclusion}
 \label{sec7}
 
This paper presented a decision-making framework for controlling an autonomous vehicle as it navigates through an unsignalized intersection crowded with pedestrians while it receives noisy observations. We improved upon pure deep reinforcement learning algorithm by including 1) a belief update LSTM model that obtains the perceived state of the environment in the case of imperfect observations, and 2) a collision detection module which masks the unsafe actions of the agent in the case of a future collision based on the future trajectories of the ego vehicle and pedestrians. We investigated the effectiveness of the belief update LSTM model and the collision detection module in improving the AVs navigating task among pedestrians. We empirically demonstrated that the proposed approach outperforms the standard rule-based agent provided by CARLA denoting that it learns a policy capable of navigating safely around pedestrians within a limited time frame while following the traffic rules with zero collision rate. We also consider how these methods trained on one intersection topology perform on another, different topology. The results reveal that our methods maintain almost the same performance.

This work assumes a Gaussian distribution of noise for pedestrian movements. Although this assumption is reasonable, there may be rare instances, such as a child darting into the street, which is not well modeled by a Gaussian and could still result in a collision. Moreover, even though our SRL method performed well, it still did not completely traverse the intersection in $8-9\%$ of the trials. Future work will focus on improving this result.  
 

\bibliographystyle{IEEEtran}
\bibliography{IEEEabrv,Bibliography}

\begin{IEEEbiography}[{\includegraphics[width=1in,height=1.25in,clip,keepaspectratio]{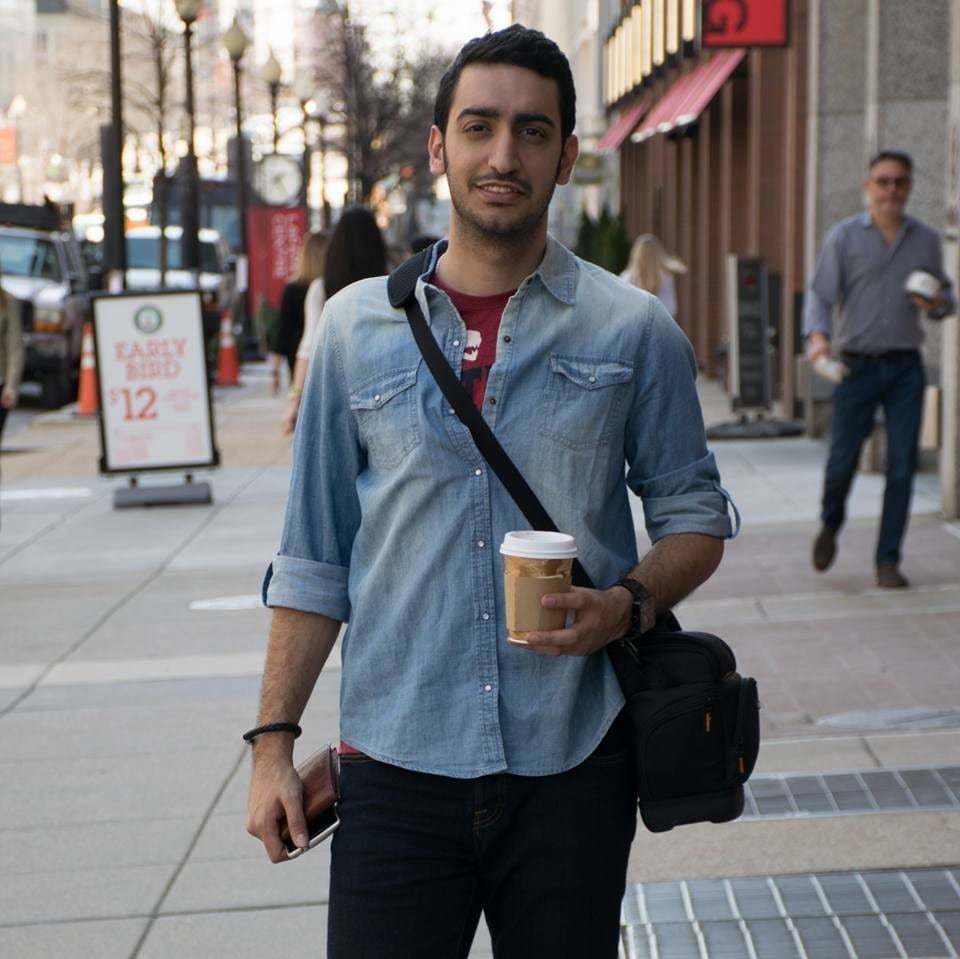}}]{Kasra Mokhtari} is a graduate research assistant in the mechanical engineering department at the Pennsylvania State University. He is currently pursuing a PhD in mechanical engineering working on incorporating social information into an autonomous vehicle decions-making process and control under Dr. Alan Wagner. He also holds a master’s degree in mechanical engineering from  the Pennsylvania State University and a bachelor’s degree in mechanical engineering from Shiraz University, Iran focusing on design and manufacturing of a Cable Robot. His areas of interest include robotics, autonomous vehicle, decision-making, reinforcement learning, and deep learning.
\end{IEEEbiography}
\begin{IEEEbiography}[{\includegraphics[width=1in,height=1.25in,clip,keepaspectratio]{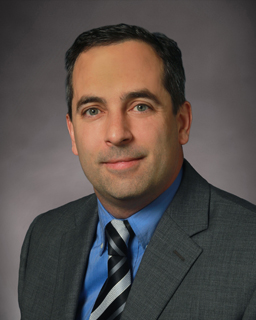}}]{Alan R. Wagner} received his Ph.D. in computer science from Georgia Institute of Technology. He also holds a master’s degree in computer science from Boston University and a bachelor’s degree in psychology from Northwestern University. His research focuses on human-robot interaction, ethics, and aerial vehicle control and teaming. He is particularly interested in methods for understanding how, when, and why people trust UAVs and developing UAVs which decide how, when and why to trust people. Application areas for these interests range from military to search and rescue. His research has won several awards including being selected for by the Air Force Young Investigator Program and being described as the 13th most important invention of 2010 by Time Magazine. 
\end{IEEEbiography}
\vfill
\end{document}